\documentclass[runningheads]{llncs}
\usepackage{graphicx}
\usepackage{multirow}
\usepackage{tikz}
\usepackage{comment}
\usepackage{amsmath,amssymb} 
\usepackage{bbm}
\usepackage{graphicx}
\usepackage{color}

\usepackage[accsupp]{axessibility}  

\begin{document}
\pagestyle{headings}
\mainmatter
\def\ECCVSubNumber{100}  

\title{A Simple Transformer-Based Model for Ego4D Natural Language Queries Challenge} 

\titlerunning{Simple Transformer for Ego4D NLQ} 
\authorrunning{Mo et al.} 
\author{Anonymous ECCV submission}
\institute{Paper ID \ECCVSubNumber}
\author{Sicheng Mo, Fangzhou Mu, Yin Li}
\institute{University of Wisconsin-Madison}

\maketitle

\begin{abstract}\vspace{-0.5em}

This report describes Badgers@UW-Madison, our submission to the Ego4D Natural Language Queries (NLQ) Challenge. Our solution inherits the point-based event representation from our prior work on temporal action localization~\cite{zhang2022actionformer}, and develops a Transformer-based model for video grounding. Further, our solution integrates several strong video features including SlowFast~\cite{SlowFast}, Omnivore~\cite{Omnivore} and EgoVLP~\cite{kevin2022egovlp}. Without bells and whistles, our submission based on a single model achieves $12.64\% $ Mean R@1 and is ranked 2$nd$ on the public leaderboard. Meanwhile, our method garners 28.45\% (18.03\%) R@5 at tIoU=0.3 (0.5), surpassing the top-ranked solution by up to 5.5 absolute percentage points.


\end{abstract}

\section{Introduction}\vspace{-0.5em}

 Given an untrimmed video and a text query, the Ego4D Natural Language Queries (NLQ) task seeks to localize the temporal window within the video where the answer to the query is evident~\cite{Ego4d_Grauman_2022_CVPR}. NLQ provides a first step towards searching our egocentric visual experience using natural language, and thus opens up the opportunity for a new generation of AI assistants. Similar to text grounding in videos (a.k.a. video grounding~\cite{lan2021survey}), NLQ requires the understanding and reasoning of the text query and video content, yet under the interference of ego-motion within egocentric video, thereby posing additional challenges. 

Prior solutions to NLQ adopted sophisticated model design for video grounding~\cite{kevin2022egovlp,liu2022reler}. Our solution instead explores a minimalist design based on Transformer~\cite{vaswani2017attention}. Specifically, we re-purpose ActionFormer, our prior work on temporal action localization~\cite{zhang2022actionformer} for video grounding. The resulting model considers every moment in the video as an event candidate, measures their relevance to the text query, and regresses the event boundaries from foreground moments. 

We show that our simple model works surprisingly well on the Ego4D NLQ task. Further, we experiment with different video features including SlowFast~\cite{SlowFast}, Omnivore~\cite{Omnivore} and EgoVLP~\cite{kevin2022egovlp}, and integrate them into our model to further boost the performance. Our final submission based on a single model achieves $12.64\% $ Mean R@1 and is ranked 2$nd$ on the public leaderboard. Meanwhile, our solution presents the highest R@5 results on the leaderboard and surpasses the top-ranked solution by up to 5.5 absolute percentage points.



\section{Approach}\vspace{-0.5em}

Our method represents a video as a sequence of feature vectors, where each vector is derived from a short clip using pre-trained video backbones. Our Transformer-based model further processes this sequence of video features, computes their similarity to an embedding of the text query, and decodes temporal event segments from moments with high similarity scores. 



\begin{figure}[!t]
\centering
\includegraphics[width=0.85\linewidth]{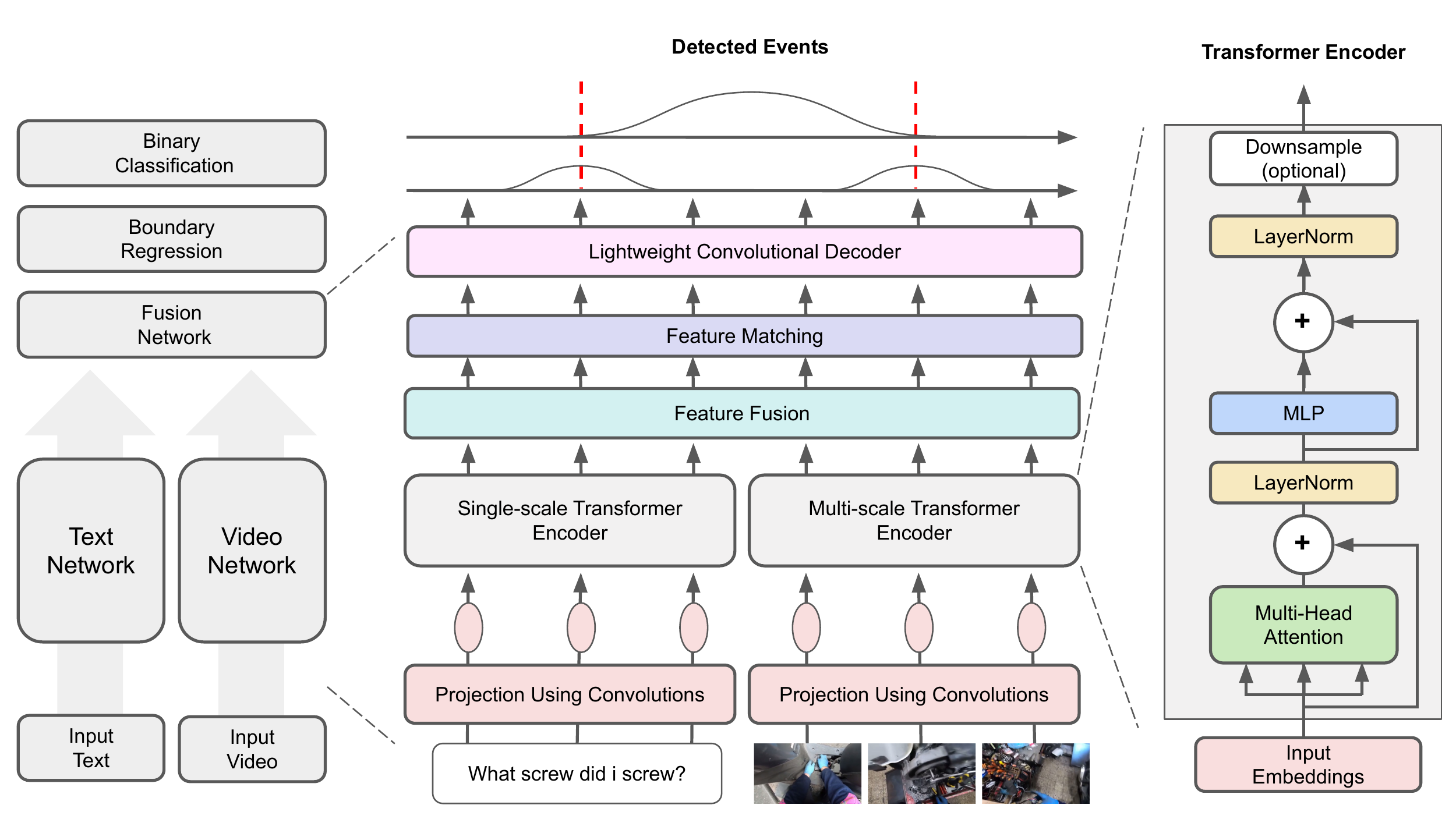} \vspace{-1.5em}
\caption{\textbf{Overview of our model.} Our model consists of a video branch and a text branch, both based on transformers. A similarity score is computed between every video and text embeddings. Those with high similarity are fused and undergo further processing for event boundary regression.}
\vspace{-1.5em}
\label{fig:overview}
\end{figure}

Figure~\ref{fig:overview} illustrates the design of our model. Our model borrows its video encoder from Actionformer~\cite{zhang2022actionformer} and runs an additional Transformer-based text encoder to obtain an embedding of the text query. The video and text embeddings are subsequently fused and further examined by shared classification and regression heads as in ActionFormer, yielding event predictions. 
\smallskip

\noindent \textbf{Input Video/Text Representations.} We use the official release of Slowfast~\cite{SlowFast} and Omnivore~\cite{Omnivore} features as our video representation.  
We additionally include video features from EgoVLP~\cite{kevin2022egovlp}, a video-language pre-training method tailored to egocentric videos. We fuse the features via simple concatenation and feed them into the video branch of our grounding model. For text queries, we extract token-wise embeddings using CLIP~\cite{radford2021learning} and feed them into the text branch.\smallskip




\noindent \textbf{Video Encoder.} Our video encoder adopts the same design as our prior work~\cite{zhang2022actionformer}. It consists of two embedding convolutions followed by seven Transformer blocks with local self-attention. The last five blocks additionally perform 2x down-sampling with depth-wise strided convolutions, resulting in a feature pyramid of six levels. More details can be found in~\cite{zhang2022actionformer}. \smallskip

\noindent \textbf{Text Encoder.} Our text encoder is also a Transformer network, consisting of a linear feature projection layer followed by multiple Transformer layers. The output text embeddings share the same length as the input. \smallskip


\noindent \textbf{Feature Fusion.} We inject textual information into video embeddings at the top of the encoders using Adaptive Attention Normalization (AdaAttN)~\cite{liu2021adaattn}. Intuitively, AdaAttN aligns the distribution of video / text features weighted by their attention scores. Consequently, the modulated video features are made aware of the text query, hence can inform event classification and offset regression.\smallskip

\noindent \textbf{Classification and Regression Heads.} The heads adopt the same convolutional design and are shared across pyramid levels, similar to those in~\cite{zhang2022actionformer}. The classification head outputs a binary score for each point on the pyramid, whereas the regression head predicts the distances from a foreground point to the event's onset and offset.\smallskip


\noindent \textbf{Loss Function.} Our loss function is a summation of three terms, including the same classification and regression losses from ActionFormer~\cite{zhang2022actionformer} and an additional, novel NCE loss inspired by the winning entry of 1$st$ Ego4D NLQ Challenge. To evaluate the NCE loss, we label a moment as positive if it lies inside a ground-truth event and negative otherwise. We pair the moments with the text embedding to form positive and negative training pairs. We apply the NCE loss on the video embeddings \textit{immediately after feature fusion} to encourage early separation of positive and negative moments.\smallskip

\noindent \textbf{Implementation Details.} Our model maintains an embedding dimension of 512 throughout the network. The Transformer layers employ 16 heads for self-attention and AdaAttN. The model is trained using the AdamW optimizer~\cite{loshchilov2017decoupled} for 9 epochs. We use a mini-batch size of 16 and a learning rate of 1e-3 with linear warm-up and consine decay. At inference time, the model outputs at most 2,000 event predictions for each video. The initial predictions are further combined and refined with SoftNMS~\cite{Bodla_2017_ICCV}, yielding final predictions subject to evaluation.
\begin{table}[t]
\centering
\caption{\textbf{Results on the Ego4D NLQ dataset.} All results on the test set are evaluated on the submission server. Results from concurrent works are {\color[HTML]{680100}colored}.}\vspace{-0.5em}
\resizebox{0.95\textwidth}{!}
{
\begin{tabular}{ccc|ccc|cc}
\hline
                        &                         &                                 & \multicolumn{3}{c|}{R@1}                                    & \multicolumn{2}{c}{R@5}                                     \\ \cline{4-8} 
\multirow{-2}{*}{Model} & \multirow{-2}{*}{Split} & \multirow{-2}{*}{Video Feature} & IoU = 0.3                    & IoU = 0.5     & ~mean~                 & IoU = 0.3                    & IoU = 0.5                  \\ \hline
EgoVLP                  & Val                     & EgoVLP                          & 10.84                        & 6.81          & 8.83              & 18.84                        & 13.45                        \\
ReLER@ZJU-Alibaba       & Val                     & SlowFast                        & 10.79                        & 6.74          &8.77               & 13.19                        & 8.85                         \\
ReLER@ZJU-Alibaba       & Val                     & Omnivore                        & 10.74                        & 6.87          &8.81               & 13.47                        & 8.72                         \\
ReLER@ZJU-Alibaba       & Val                     & Ensemble                        & 11.33                        & 7.05          &9.19               & 14.77                        & 8.98                         \\
Ours                    & Val                     & SlowFast                        & 9.96                         & 6.63          &8.30               & 25.86                        & 16.80                        \\
Ours                    & Val                     & Omnivore                        & 11.46                      & 7.28            &9.37             & 28.42                        & 17.71                        \\
Ours                    & Val                     & EgoVLP                          & 12.03                        & 7.15          &9.59               & 28.34                        & 17.71                        \\
Ours                    & Val                     & Fused                           & 15.72                        & 10.12         &12.92               & 34.64                        & 23.64                        \\ \hline \hline
EgoVLP                  & Test                    & EgoVLP                          & 10.46                        & 6.24          &8.35               & 16.76                        & 11.29                        \\
ReLER@ZJU-Alibaba       & Test                    & Ensemble                        & 12.89                        & 8.14           &10.52              & 15.41                        & 9.94                         \\
CONE (3rd)       & Test                    &                                 & {\color[HTML]{680100} 15.26} & {\color[HTML]{680100} 9.24}  &{\color[HTML]{680100} 12.25}  & {\color[HTML]{680100} 26.42} & {\color[HTML]{680100} 16.51} \\
Red Panda (1st)       & Test                    &                                 & {\color[HTML]{680100} \textbf{16.46}} & {\color[HTML]{680100} \textbf{10.06}} &{\color[HTML]{680100} \textbf{13.26}} & {\color[HTML]{680100} 22.95} & {\color[HTML]{680100} 16.11} \\
Ours (2nd)                    & Test                    & Fused                           & 15.71                        & 9.57         &12.64                & \textbf{28.45}                        & \textbf{18.03}                        \\ \hline
\end{tabular}
}\vspace{0.4em}
\label{tab:results}\vspace{-2em}
\end{table}

\section{Experiments and Results}\vspace{-0.5em}

We now describe our experiments on the Ego4D NLQ dataset. We first present an ablation study of our method using different video features, then compare our results to other methods on the leaderboard.\smallskip 

\noindent \textbf{Dataset and Evaluation Metrics.}
The Ego4D NLQ dataset contains 227 hours of videos with 5.9K clips and 22.5K pairs of events and text queries. The video clips last either 8 or 20 minutes. Text queries are generated from 13 pre-defined templates. We follow the official train/val/test splits in our experiments. We train our model on the train split when reporting results on the val split, and train on the combined train/val splits when reporting results on the test split. We report Recall@1 (R@1) and Recall@5 (R@5) at tIoU thresholds 0.3 and 0.5.\smallskip


\noindent \textbf{Results.} Table~\ref{tab:results} summarizes our results. On the val split, our method is on par with the champion from last challenge (ReLER@ZJU-Alibaba~\cite{liu2022reler}) in R@1 scores but is better in R@5. Among the three types of features (SlowFast, Omnivore and EgoVLP), EgoVLP alone yields the best results thanks to the in-domain pre-training. An ensemble of all three features further boosts all metrics by a significant amount. On the test split, our model ultimately achieves $12.64\%$ Mean R@1, 2.12\% absolute percentage points higher than the last champion~\cite{liu2022reler}. Our Mean R@1 score lags behind the top-ranked solution by a tiny gap of 0.62\%. Meanwhile, our R@5 scores beat all entries on the leaderboard, and in particular, surpasses the top-ranked solution by up to 5.5 absolute percentage points.\smallskip

\begin{figure}[!t]
\centering
\includegraphics[width=0.9\linewidth]{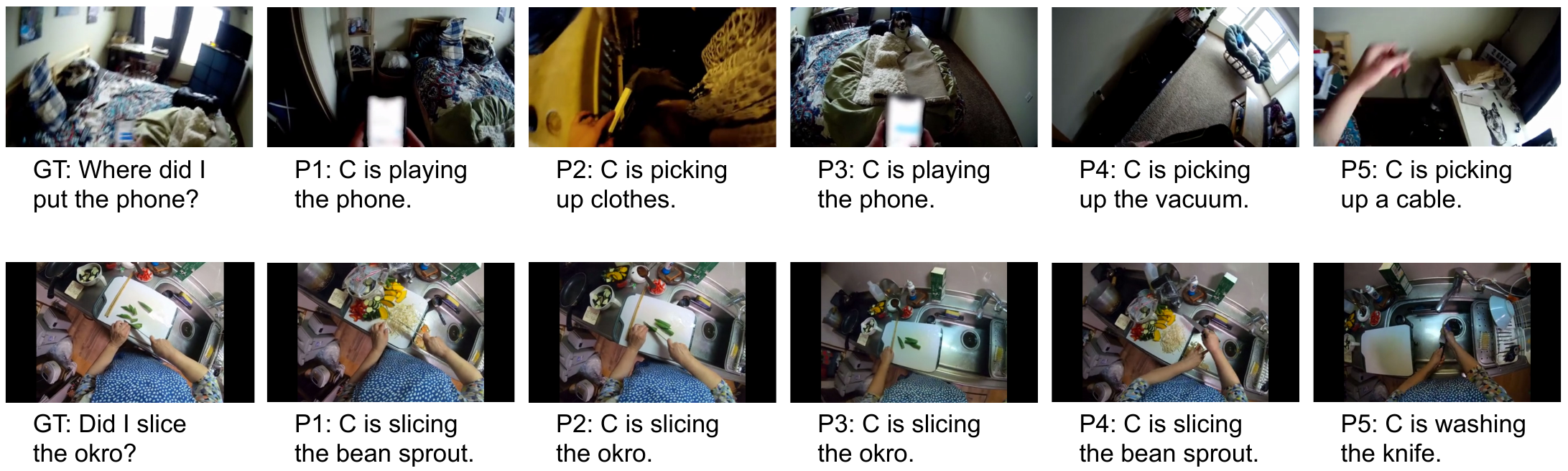} \vspace{-1em}
\caption{\textbf{Failure cases of our model.} `GT' are input text queries and ground-truth events displayed as key frames. `P\#' are the top-5 predictions with text descriptions derived from video frames within the prediction window.}
\vspace{-1.5em}
\label{fig:failure}
\end{figure}

\noindent \textbf{Limitations and Discussion.} Some failure cases of our model are shown in Figure~\ref{fig:failure}. A common failure mode is that our model sticks to the moments where the object of interest is present, yet failed to localize the moment from which the answer to the question can be deduced. We conjecture that explicit reasoning about the text queries might be necessary to avoid those errors. Another interesting observation is that the R@5 scores of our model are considerably higher than other methods, suggesting that learning stronger matching / classification heads is a promising future direction. 

\section{Conclusion}

In this report, we presented our solution to the Ego4D NLQ task. Our solution combines a variant of a latest temporal action localization backbone with strong video features, resulting in impressive empirical results on the public learderboard while maintaining a minimalist model design. We hope our model design and results can shed light on video grounding and egocentric vision.


%
%
\bibliographystyle{main}
\bibliography{main}
\end{document}